\documentclass[10pt,twocolumn,letterpaper]{article}

\usepackage[pagenumbers]{cvpr}

\usepackage{graphicx}
\usepackage{amsmath,amssymb}
\usepackage{booktabs}
\usepackage{multirow}
\usepackage{xcolor}
\usepackage{colortbl}
\usepackage{microtype}

\definecolor{cvprblue}{rgb}{0.21,0.49,0.74}
\usepackage[pagebackref,breaklinks,colorlinks,allcolors=cvprblue]{hyperref}

\def\paperID{}
\def\confName{Preprint}
\def\confYear{2026}

\definecolor{vlmcolor}{HTML}{E65100}
\definecolor{gpucolor}{HTML}{1565C0}
\definecolor{deltaup}{HTML}{B71C1C}

\title{Can a Teenager Fool an AI?\\ %
Evaluating Low-Cost Cosmetic Attacks on Age Estimation Systems}

\author{
Xingyu Shen$^{1,3*}$, Tommy Duong$^{2*}$, Xiaodong An$^{1,4}$, Zengqi Zhao$^{5}$, Zebang Hu$^{1}$, \\
Haoyu Hu$^{6}$, Ziyou Wang$^{1}$, Finn Guo$^{3}$, Simiao Ren$^{1\dagger}$
\vspace{2mm} \\
$^{1}$Reality Inc. \quad $^{2}$UC Berkeley \quad $^{3}$Duke University \\
$^{4}$Georgia Tech \quad $^{5}$UNC Chapel Hill \quad $^{6}$UC San Diego \\
\vspace{1mm} \\
{\tt\small alex@get-reality.com \quad tommy-duong0@berkeley.edu \quad xan37@gatech.edu} \\
{\tt\small zengqi@ad.unc.edu \quad jasperh.work@gmail.com \quad hah034@ucsd.edu} \\
{\tt\small ziw147@ucsd.edu \quad xg101@duke.edu \quad benren@scam.ai} \\
\vspace{1mm} \\
\small{$^*$Equal contribution \quad $^\dagger$Corresponding author}
}

\begin{document}
\maketitle
{\renewcommand{\thefootnote}{}\footnote{Authorship of this paper does not imply employment by or affiliation with Scam.AI (Reality Inc.); some authors are independent researchers who collaborated on this work.}\addtocounter{footnote}{-1}}

\begin{abstract}
Age estimation systems are increasingly deployed as gatekeepers for age-restricted online content, yet their robustness to cosmetic modifications has not been systematically evaluated.
We investigate whether simple, household-accessible cosmetic changes---beard, grey hair, makeup, and simulated wrinkles---can cause AI age estimators to classify minors as adults.
To study this threat at scale without ethical concerns, we \emph{simulate} the physical attacks on 329 facial images of individuals aged 10--21 using a VLM image editor (Gemini~2.5 Flash Image).
We then evaluate eight models from our prior benchmark: five specialized architectures (MiVOLO, Custom-Best, Herosan, MiViaLab, DEX) and three vision-language models (Gemini~3 Flash, Gemini~2.5 Flash, GPT-5-Nano).
We introduce the \textbf{Attack Conversion Rate (ACR)}: the fraction of images that a model predicted as minor at baseline which flip to adult after attack---a population-agnostic metric that does not depend on the ratio of minors to adults in the test set.
Our results reveal that (1)~a synthetic beard alone achieves 28--69\% ACR across all eight models; (2)~combining all four attacks shifts predicted age by $+7.7$ years on average (across all 329 subjects) and reaches up to 83\% ACR; and (3)~VLMs exhibit lower ACR (59--71\%) than specialized models (63--83\%) under the full attack, though ACR ranges overlap and the difference is not statistically tested.
These findings highlight a critical vulnerability in deployed age verification pipelines and call for adversarial robustness evaluation as a mandatory criterion for model selection.
\end{abstract}

\section{Introduction}
\label{sec:intro}

\begin{figure*}[t]
  \centering
  \includegraphics[width=\linewidth]{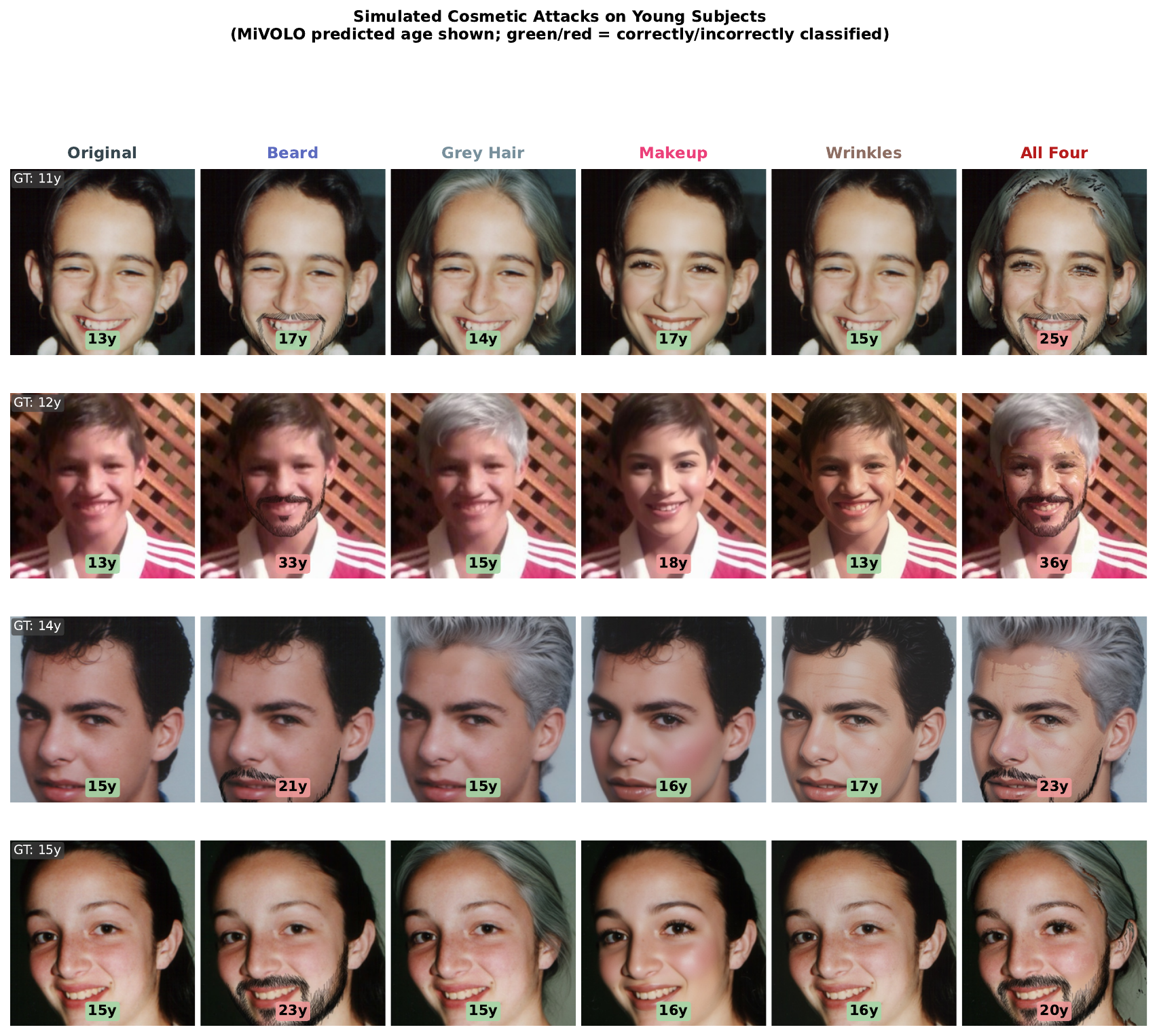}
  \caption{Simulated cosmetic attacks on four subjects (ground-truth ages 11--15). Predicted age shown per condition (MiVOLO). Green badge = correctly classified as minor; red badge = bypassed (predicted adult). The all-four combination reliably fools the model across all subjects.}
  \label{fig:examples}
\end{figure*}

Age verification is a cornerstone of online child safety regulation.
The UK Online Safety Act~\cite{osa2023}, the EU Digital Services Act~\cite{eu2022dsa}, and proposed US legislation~\cite{kosa2022} all mandate that platforms restrict minors from harmful content, with facial age estimation emerging as a scalable, low-friction mechanism for enforcement~\cite{ofcom2024ageverification}.
Yet survey data from Ofcom~\cite{ofcom2023childrenonline} reveals that 44\% of UK 8--17-year-olds have lied about their age online, and applying cosmetics to appear older is a plausible low-effort strategy that requires no technical knowledge.
Figure~\ref{fig:examples} illustrates how four simple cosmetic modifications---applied via AI image editing---can systematically fool a state-of-the-art age estimator.

Despite this threat landscape, the robustness of AI age estimators to simple cosmetic modifications has received virtually no systematic study.
Robustness gaps have been identified across adjacent AI content analysis domains---deepfake detectors substantially degrade under simple super-resolution post-processing~\cite{ren2025deepfakereality}, and AI-generated image detectors fail to generalize across deployment settings despite strong in-distribution performance~\cite{ren2026aidetect}---yet age estimation systems have not been subjected to analogous adversarial scrutiny.
Prior adversarial work on face-related systems focuses on digital perturbations~\cite{goodfellow2015explaining}, adversarial patches~\cite{brown2017adversarial}, or generative makeup attacks~\cite{yin2021advmakeup,hu2022sintm} designed to defeat \emph{face recognition}---not age estimation.
Age estimation models, whether specialized convolutional networks or modern vision-language models (VLMs), are trained primarily on age-labeled datasets and are evaluated solely on mean absolute error (MAE) without any adversarial robustness assessment.

\paragraph{Our contribution.}
We present the first systematic study of \emph{physical cosmetic attacks} on AI age estimation.
Critically, our attacks differ from all prior adversarial work on face analysis in a fundamental way: existing methods require either gradient access to a target model, adversarially optimized digital perturbations, or prompt injection---techniques that demand algorithmic sophistication and are unavailable to a typical minor.
Our attacks instead simulate what a determined teenager could apply before a camera in under ten minutes using drugstore cosmetics: a beard, grey hair, makeup, and simulated wrinkles.
We evaluate these attacks and all 15 non-empty subsets thereof across 329 images and 8 models, using a strict simulation protocol (VLM image editing) that avoids any ethical concerns around modifying real subjects.

\begin{itemize}
  \item \textbf{First systematic physical cosmetic attack study on age estimation}: To the best of our knowledge, no prior work has studied physical cosmetic modifications as a deliberate attack on age estimation.
  Prior adversarial makeup work~\cite{yin2021advmakeup,hu2022sintm,shamshad2023clip2protect} targets face \emph{recognition} (identity bypass) via gradient optimization.
  Passive accuracy audits~\cite{chen2014cosmetics,anda2020underage} have observed that makeup degrades age estimation accuracy, but do not frame it as intentional evasion, do not evaluate beard or hair modifications, and do not measure bypass rates.

  \item \textbf{Cross-paradigm robustness benchmark}: We evaluate 5 specialized CV architectures and 3 zero-shot VLMs from our prior benchmark~\cite{ren2025benchmark} on the same attack suite, enabling the first direct comparison of robustness across model paradigms under a shared real-world threat model.
\end{itemize}

Our results are striking: a synthetic beard alone achieves 28--69\% Attack Conversion Rate (ACR)---the fraction of model-flagged minors that are reclassified as adults after attack.
Combining all four attacks yields a mean age shift of $+7.7$ years and up to 83\% ACR.
VLMs exhibit lower ACR (59--71\%) than specialized models (63--83\%), though ranges overlap.
These findings reveal that the technology now being deployed as global age-gating infrastructure is vulnerable to low-cost, no-expertise circumvention.

The remainder of this paper is organized as follows: \Cref{sec:related} surveys related work, \Cref{sec:threat} describes our threat model, \Cref{sec:attacks} details the attack simulation methodology, \Cref{sec:models} introduces the evaluated models, \Cref{sec:experiments} presents results, and \Cref{sec:conclusion} concludes.

\section{Related Work}
\label{sec:related}

\paragraph{Age estimation.}
Facial age estimation has evolved from handcrafted feature methods~\cite{guo2009agewild} through convolutional neural networks (DEX~\cite{rothe2015dex}, C3AE~\cite{zhang2019c3ae}) to modern Vision Transformer architectures such as MiVOLO~\cite{kuprashevich2024mivolo}, which achieves state-of-the-art performance by jointly reasoning over face and body features.
Our prior benchmark~\cite{ren2025benchmark} evaluated 34 models on 8 standard datasets and demonstrated that zero-shot VLMs now rival or surpass task-specific architectures in MAE.
Paplham and Franc~\cite{paplhjak2024calltoreflect} conduct a complementary analysis of evaluation practices for age estimation, highlighting dataset biases and the need for standardized benchmarking --- concerns that motivate our use of multiple datasets and a population-agnostic metric.
However, that study evaluated models only on unmodified test images; the present work evaluates the same models under adversarial cosmetic modifications.
Two prior studies have observed that cosmetics passively degrade age estimation accuracy: Chen et al.~\cite{chen2014cosmetics} measured MAE increases of up to 5.84 years under real and synthetically retouched makeup, and Anda et al.~\cite{anda2020underage} identified makeup as one of several factors degrading accuracy on minor images.
Neither study frames cosmetics as a deliberate evasion strategy, evaluates beard or hair-color modifications, nor measures adversarial bypass rates.
The present work is the first to study all four cosmetic modifications as an intentional, combined attack targeting the age gate, with attack conversion rate as the central metric.

\paragraph{Adversarial attacks on face analysis.}
Digital adversarial attacks on face recognition~\cite{goodfellow2015explaining,carlini2017towards} are well-studied, but physical attacks that survive real-world capture are more relevant to our setting.
Eyeglass-based physical adversarial patches~\cite{sharif2016accessorize} were among the earliest physical attacks on face recognition.
More recently, adversarial makeup approaches---Adv-Makeup~\cite{yin2021advmakeup}, AMT-GAN~\cite{hu2022sintm}, and CLIP2Protect~\cite{shamshad2023clip2protect}---apply natural-looking cosmetics that are optimized via gradient descent to fool face recognition pipelines.
DiffAM~\cite{sun2024diffam} uses diffusion models to generate transferable adversarial makeup.
Beyond adversarially optimized perturbations, simple real-world transformations have also been shown to undermine face forensics models: super-resolution post-processing applied to deepfake imagery causes detector accuracy to approach random chance~\cite{ren2025deepfakereality}, illustrating the broader distributional fragility of face analysis systems.
Critically, all of these attacks are optimized for \emph{face recognition} (identity bypass) rather than age estimation, and they require access to the target model's gradients or outputs---capabilities unavailable to a non-technical actor.
Our work differs in two fundamental respects: (1)~we study age estimation, not identity recognition; and (2)~the attacks we simulate require \emph{no model access, no gradient computation, and no algorithmic knowledge}---they represent cosmetic changes that any teenager could apply with items available at a drugstore.
This threat model is qualitatively different from prior adversarial work, and it is the relevant one for the age verification deployment context.

\paragraph{Age verification policy.}
Regulatory pressure to enforce age limits online has intensified dramatically.
In the United States, more than 20 states have enacted age verification statutes for adult content and social media platforms, including early movers such as Louisiana (2023), Utah, Virginia, Texas, and Florida (2025)~\cite{kosa2022}.
The UK Online Safety Act~\cite{osa2023} entered enforcement in 2024 and explicitly permits facial age estimation as a compliant verification mechanism~\cite{ofcom2024ageverification}.
Australia's Online Safety Amendment (2024) bans users under 16 from social media platforms and mandates age verification at scale.
The EU Digital Services Act~\cite{eu2022dsa} requires age-appropriate design, with member states including France, Spain, and Italy actively piloting age estimation technologies.
Across these frameworks, facial age estimation is emerging as the preferred low-friction solution, replacing document-upload and credit-card checks that create friction for legitimate adult users.
Yet none of these regulatory frameworks specify minimum adversarial robustness requirements for approved age estimation systems.
The NIST FATE project~\cite{nist2019fate} evaluates demographic accuracy gaps but does not consider adversarial inputs.
Our work directly addresses this regulatory blind spot: as facial age estimation becomes the infrastructure of online age gating at global scale, its vulnerability to simple cosmetic bypass is a first-order policy concern.

\paragraph{Age progression and makeup simulation.}
VLM-based image editing (InstructPix2Pix~\cite{brooks2023instructpix2pix}, Gemini~2.5 Flash Image~\cite{comanici2025gemini25}) can apply realistic appearance changes from text prompts, enabling our large-scale simulation without requiring physical photography.
GANs have been used for age progression/regression~\cite{zhang2017utkface,wang2018face} and aging simulation~\cite{or2020lifespan}, but these produce full-face transformations that are visually conspicuous.
Our simulation targets the specific cosmetic items that minors realistically apply: beard, grey hair, makeup, and wrinkles.

\section{Threat Model}
\label{sec:threat}

\paragraph{Adversary profile.}
We model an adversary who is a minor (age $< 18$) attempting to bypass an online age verification gate.
The adversary has no technical knowledge and no access to the target model or its architecture.
Their resources are limited to commonly available cosmetic supplies: a fake beard, hair dye or grey wig, foundation and blush (makeup), and theatrical wrinkle cream.
These items are sold at costume shops and online retailers for under \$30 and require no specialized skill to apply.
This models a plausible, low-effort strategy that a motivated teenager could readily employ, as opposed to digital adversarial attacks~\cite{goodfellow2015explaining} or gradient-optimized makeup~\cite{yin2021advmakeup} that require technical expertise.

\paragraph{Attack surface.}
The victim is an age estimation system that takes a face photograph as input and returns a predicted age or an adult/minor classification.
Such systems are deployed as selfie-based age gates on platforms including adult content sites, alcohol retailers, gambling services, and social media platforms.
The system may be a specialized age regression model or a VLM-based system; we evaluate both.

\paragraph{Success criterion and metrics.}
A bypass is successful if the system predicts $\hat{y} \geq 18$ for a subject with true age $< 18$.
We report two complementary metrics:

\noindent\textbf{Mean age shift} ($\Delta\bar{y}$): the average change in predicted age across all subjects,
\begin{equation}
  \Delta\bar{y} = \frac{1}{n}\sum_{i=1}^{n}\bigl(\hat{y}_i^{\text{atk}} - \hat{y}_i^{\text{base}}\bigr),
\end{equation}
measuring how many years older the model perceives the subject after applying the cosmetic attack.

\noindent\textbf{Attack Conversion Rate (ACR)}: among images where the baseline model predicted ``minor'' ($\hat{y}_i^{\text{base}} < 18$), what fraction flip to ``adult'' ($\hat{y}_i^{\text{atk}} \geq 18$) after the attack:
\begin{equation}
  \text{ACR} = \frac{\bigl|\{i : \hat{y}_i^{\text{base}} < 18,\; \hat{y}_i^{\text{atk}} \geq 18\}\bigr|}
                    {\bigl|\{i : \hat{y}_i^{\text{base}} < 18\}\bigr|} \times 100\%.
\end{equation}
ACR is \textbf{population-agnostic}: its denominator depends only on the model's own baseline predictions, not on the ground-truth age distribution of the test set.
This makes ACR directly comparable across studies with different test-set demographics.
An ACR of 50\% means the attack converts half of the cases the model was predicting as minor; an ACR of 100\% means the model predicts adult for every subject it previously classified as minor.
Note that the denominator includes all images the model predicted as minor regardless of ground-truth age; a model that misclassifies some adults as minor will include those in its denominator.

\paragraph{Scope.}
Our threat model explicitly excludes: (1)~attacks requiring digital post-processing of the photograph; (2)~attacks that are visually non-natural and would be detected by human review; (3)~gradient-based attacks that require model access.
We focus exclusively on cosmetic modifications that are invisible to a casual human observer---the same cosmetics a teenager might apply for a night out.

\paragraph{Ethical considerations.}
We did \emph{not} conduct this study by physically applying cosmetics to minors and photographing them.
Instead, we simulate all attacks digitally by prompting a VLM image editor to add the relevant cosmetic features to images drawn from publicly available research datasets (UTKFace~\cite{zhang2017utkface}, IMDB-WIKI~\cite{rothe2018imdbwiki}, MORPH~\cite{ricanek2006morph}, AFAD~\cite{niu2016afad}, CACD~\cite{chen2015cacd}, FG-NET~\cite{panis2016fgnet}, APPA-REAL~\cite{agustsson2017apparent}, AgeDB~\cite{moschoglou2017agedb}).
This simulation approach allows large-scale, reproducible evaluation without any human subjects research.
All images are from publicly released research datasets with standard research-use licenses.

\section{Attack Simulation Methodology}
\label{sec:attacks}

\subsection{Image Dataset}

We construct a test set of 329 face images drawn from eight standard age estimation datasets~\cite{zhang2017utkface,rothe2018imdbwiki,ricanek2006morph,niu2016afad,chen2015cacd,panis2016fgnet,agustsson2017apparent,moschoglou2017agedb} (same datasets as our benchmark~\cite{ren2025benchmark}), restricted to individuals with ground-truth ages 10--21.
Images were filtered to ensure clear frontal face visibility, adequate resolution ($\geq 120 \times 120$ pixels), and RGB color (no grayscale).
Of the 329 images, 168 are true minors (age $< 18$, ages 10--17) and 161 are young adults (ages 18--21); together they span the decision boundary relevant to age gating.
We use all 329 images for computing mean age shift and MAE metrics. ACR is computed per-model over the subset of images that the baseline model predicted as minor ($\hat{y} < 18$), as defined in \Cref{sec:threat}.

\subsection{VLM Image Editor}

We simulate four cosmetic attack types using \textbf{Gemini~2.5 Flash Image}~\cite{comanici2025gemini25}, a state-of-the-art VLM image editor capable of applying realistic local modifications from text instructions.
For each attack type, we use a standardized prompt that specifies the desired cosmetic change while instructing the model to preserve other facial features.
Table~\ref{tab:prompts} lists the prompts used.

\begin{table}[t]
\centering
\small
\caption{Prompts used for VLM-based attack simulation. All prompts begin with ``Apply the following modification to the face in this image:''.}
\label{tab:prompts}
\begin{tabular}{lp{5.0cm}}
\toprule
Attack & Prompt suffix \\
\midrule
Beard       & Add a realistic full beard with slight gray mixed in. Keep all other features unchanged. \\
Grey hair   & Turn the hair silver-grey as if naturally aging. Preserve the hairstyle and face. \\
Makeup      & Apply mature foundation, subtle blush, and light contouring that adds apparent age. \\
Wrinkles    & Add natural forehead and eye-corner wrinkles consistent with a 35--50 year old. \\
\bottomrule
\end{tabular}
\end{table}

The output images are $1024 \times 1024$ pixels; we downscale to match each original image's resolution using area interpolation (\texttt{cv2.INTER\_AREA}) before inference.
Images where the VLM failed to produce a valid output (e.g., content policy refusal, face not detected in output) are excluded.
After filtering, we obtain 307--326 images per single attack type (beard: 307, grey hair: 326, makeup: 314, wrinkles: 315).

\subsection{Combination Attacks}

To study synergistic effects, we generate all 15 non-empty subsets of the four attack types (4 singles, 6 pairs, 4 triples, 1 full combo).
Na\"ively applying attacks sequentially would degrade image quality (each generation step adds artifacts).
Instead, we use a \textbf{priority-weighted pixel-delta} blending strategy:

\begin{enumerate}
  \item Apply each single attack independently, obtaining per-pixel deltas $\delta_k = I_k - I_\text{orig}$ for attack $k$.
  \item Assign a priority order: beard $>$ grey hair $>$ wrinkles $>$ makeup.
  \item For each pixel, determine the \emph{owner}: the highest-priority attack whose delta magnitude exceeds a threshold $\tau = 15$ (on 8-bit channels; chosen empirically to separate meaningful edits from compression noise).
  \item Construct the combination by applying the owner's delta: $I_\text{combo}[p] = I_\text{orig}[p] + \delta_{\text{owner}(p)}[p]$.
\end{enumerate}

This strategy preserves the natural appearance of each individual attack component without compounding generation artifacts.
Beard and wrinkle regions receive their respective textures; grey hair and makeup are blended into regions not dominated by the higher-priority attacks.
The full quad-attack set produces 286 valid images (consistent across all models); the range 286--295 applies to partial combinations (pairs and triples).
Failure rates do not appear to correlate systematically with subject age within the 10--21 range studied.

\subsection{Attack Taxonomy}

\paragraph{Beard.}
The most powerful single attack in our study.
Synthetic beards are a classic stage prop and are trivially available as costume accessories.
They cover the lower face and chin, regions that provide strong age cues to both human observers and face analysis models.

\paragraph{Grey hair.}
Hair graying is strongly correlated with age across all cultures.
Grey hair dye and wigs are inexpensive and widely available.
Our simulation models this by replacing hair pixels with a desaturated silver-grey hue while preserving texture and style.

\paragraph{Makeup.}
Adult-style foundation and contouring makeup shifts apparent age through changes in skin tone uniformity and facial shadow patterns.
This attack has an unusual interaction: while it achieves substantial ACR (29--49\% across models), it has minimal effect on mean age shift ($+0.05$~years overall), suggesting that it primarily alters predictions near the decision boundary without systematically inflating predicted age.

\paragraph{Wrinkles.}
Simulated crow's feet and forehead lines mimic the most salient aging cues.
Theatrical wrinkle creams and latex prosthetics that produce these effects are routinely used in stage makeup.

\section{Evaluated Models}
\label{sec:models}

We evaluate the eight models that represented the performance spectrum in our prior cross-paradigm benchmark~\cite{ren2025benchmark}.
For completeness, Table~\ref{tab:models} summarizes key properties, including the number of images predicted as minor at baseline ($N_{\text{base}<18}$) --- the denominator used for ACR computation.

\begin{table}[t]
\centering
\small
\caption{Evaluated age estimation models. ``Clean MAE'' is the mean absolute error on the unmodified test set (329 images). $N_{\text{base}<18}$: number of images (out of 329) where the baseline model predicted minor ($\hat{y}<18$) --- the ACR denominator. Type: \textcolor{gpucolor}{\textbf{CV}} = specialized CV architecture; \textcolor{vlmcolor}{\textbf{VLM}} = zero-shot VLM.}
\label{tab:models}
\resizebox{\columnwidth}{!}{%
\begin{tabular}{llccc}
\toprule
Model & Type & Clean MAE & $N_{\text{base}<18}$ & \% of 329 \\
\midrule
MiVOLO~\cite{kuprashevich2024mivolo}         & \textcolor{gpucolor}{\textbf{CV}} & 4.35 & 126 & 38.3\% \\
Custom-Best~\cite{ren2025benchmark}           & \textcolor{gpucolor}{\textbf{CV}} & 2.71 & 108 & 32.8\% \\
Herosan~\cite{ren2025benchmark}               & \textcolor{gpucolor}{\textbf{CV}} & 6.47 &  83 & 25.2\% \\
MiViaLab~\cite{ren2025benchmark}             & \textcolor{gpucolor}{\textbf{CV}} & 6.16 &  87 & 26.4\% \\
DEX~\cite{rothe2015dex}                       & \textcolor{gpucolor}{\textbf{CV}} & 5.98 &  57 & 17.3\% \\
\midrule
Gemini~3 Flash~\cite{google2025gemini3flash}   & \textcolor{vlmcolor}{\textbf{VLM}} & 3.05 & 117 & 35.6\% \\
Gemini~2.5 Flash~\cite{comanici2025gemini25}  & \textcolor{vlmcolor}{\textbf{VLM}} & 4.23 & 121 & 36.8\% \\
GPT-5-Nano~\cite{openai2025gpt5}             & \textcolor{vlmcolor}{\textbf{VLM}} & 3.90 &  92 & 28.0\% \\
\bottomrule
\end{tabular}%
}
\end{table}

\paragraph{Specialized (CV) models.}
MiVOLO~\cite{kuprashevich2024mivolo} is the top-performing specialized model in our benchmark; it uniquely incorporates both facial and body region features via Vision Transformers.
Custom-Best~\cite{ren2025benchmark} is a fine-tuned ensemble achieving the lowest clean MAE (2.71 years) among all eight evaluated models.
Herosan and MiViaLab are publicly available specialized models from our prior benchmark~\cite{ren2025benchmark} representing mid-range performance; both use CNN-based age regression trained on large-scale face datasets.
DEX~\cite{rothe2015dex} uses expected value estimation over discrete age bins, representing the older generation of specialized architectures.

\paragraph{Vision-language models.}
Gemini~3 Flash~\cite{google2025gemini3flash}, Gemini~2.5 Flash (Google)~\cite{comanici2025gemini25}, and GPT-5-Nano (OpenAI)~\cite{openai2025gpt5} are multimodal language models prompted with: ``\textit{How old does the person in this image appear? Reply with only a number.}''
Gemini~3 Flash was accessed during its preview period (\texttt{gemini-3-flash-preview}); results may differ from the general availability release, as also noted in the limitations section.

\paragraph{Baseline protection analysis.}
The column $N_{\text{base}<18}$ in Table~\ref{tab:models} reveals an important baseline property: the fraction of images that each model predicts as minor before any attack ranges from just 17\% (DEX) to 38\% (MiVOLO).
This means DEX already predicts ``adult'' for 82.7\% of the test set unprompted.
VLMs correctly predict minor for 28--37\% of images --- comparable to or better than most specialized models.
These baseline differences directly influence each model's susceptibility to attack: a model that rarely predicts ``minor'' cannot have a high ACR from attack.

\section{Experiments}
\label{sec:experiments}

\subsection{Setup}

All models are evaluated on 329 images (ages 10--21) as described in \Cref{sec:attacks}.
We report: (1)~mean age shift $\Delta\bar{y}$, computed over all 329 subjects; (2)~Attack Conversion Rate (ACR), computed over the $N_{\text{base}<18}$ images predicted as minor at baseline; and (3)~post-attack MAE.
Combination-attack metrics are computed over the matched image subset: 286 images for the full quad-attack (consistent across all models); 286--295 images for partial combinations (pairs and triples).
Figure~\ref{fig:examples} (Introduction) shows sample attack outputs for four representative subjects.

\subsection{Single Attack Results}

Table~\ref{tab:singles} reports results for the four single-attack conditions.
Supplementary Figure~\ref{fig:violin} shows the full distribution of age shifts per attack for CV and VLM models separately.

\begin{table}[t]
\centering
\small
\caption{Single attack results averaged across 8 models. $\Delta\bar{y}$: mean age shift (yr). ACR: Attack Conversion Rate (\%).}
\label{tab:singles}
\setlength{\tabcolsep}{3.5pt}
\resizebox{\columnwidth}{!}{%
\begin{tabular}{lrrrrrrrr}
\toprule
& \multicolumn{2}{c}{Beard} & \multicolumn{2}{c}{Grey Hair} & \multicolumn{2}{c}{Makeup} & \multicolumn{2}{c}{Wrinkles} \\
\cmidrule(lr){2-3}\cmidrule(lr){4-5}\cmidrule(lr){6-7}\cmidrule(lr){8-9}
Model & $\Delta\bar{y}$ & ACR & $\Delta\bar{y}$ & ACR & $\Delta\bar{y}$ & ACR & $\Delta\bar{y}$ & ACR \\
\midrule
MiVOLO      & +4.1 & 52 & +0.5 & 13 & $-$0.1 & 35 & +1.8 & 31 \\
Custom-Best & +4.8 & 59 & +1.2 & 26 &  +0.8  & 45 & +2.3 & 34 \\
Herosan     & +5.1 & 47 & +1.2 & 24 & $-$0.6 & 49 &  0.0 & 19 \\
MiViaLab    & +2.7 & 56 & +1.4 & 20 & $-$0.5 & 29 & +0.9 & 25 \\
DEX         & +4.8 & 69 & +0.3 & 16 & $-$1.1 & 35 & +0.7 & 23 \\
\midrule
Gemini~3 Flash    & +1.6 & 28 & +3.0 & 22 & +0.6   & 32 & +2.2 & 25 \\
Gemini~2.5 Flash  & +1.9 & 29 & +4.1 & 33 & +0.4   & 37 & +2.5 & 32 \\
GPT-5-Nano  & +3.9 & 37 & +5.2 & 29 & +0.7   & 35 & +1.4 & 22 \\
\midrule
\textit{Avg}& +3.6 & 47 & +2.1 & 23 & +0.0   & 37 & +1.5 & 27 \\
\bottomrule
\end{tabular}%
}
\end{table}

\paragraph{Beard is the dominant single attack.}
A synthetic beard achieves 28--69\% ACR depending on the model and shifts predicted age by $+1.6$ to $+5.1$ years.
Specialized models show larger mean shifts (avg.\ $+4.3$ yr) than VLMs (avg.\ $+2.5$ yr), but CV models also have a much larger ACR (avg.\ 57\% vs.\ 31\%) because specialized models are more susceptible to the facial-region cue change that beard introduces.

\paragraph{Grey hair affects VLMs more than specialized models.}
Specialized models show modest response to grey hair ($+0.3$--$+1.4$ yr, ACR 13--26\%), while VLMs exhibit a stronger effect ($+3.0$--$+5.2$ yr, ACR 22--33\%).
This asymmetry suggests that VLMs leverage hair color as a contextual age cue more heavily than specialized models, which may focus on facial texture and geometry.

\paragraph{Makeup: high ACR, near-zero mean shift.}
Makeup has essentially zero mean shift (+0.05 yr averaged across all models) yet achieves a non-trivial ACR (29--49\%).
This dissociation reveals that makeup does not inflate predicted ages systemically, but selectively pushes predictions near the decision boundary from just below 18 to just above.
This is confirmed by the bimodal delta distribution visible in Supplementary Figure~\ref{fig:violin}: 74\% of model--subject pairs show near-zero shift ($|\Delta|\leq3$ yr), while 11\% show a large upward jump (mean $+5.9$ yr); this high-ACR subgroup has a mean baseline prediction of 15.9 yr---squarely near the decision boundary---confirming the selective boundary-pushing mechanism.

\paragraph{Wrinkles are effective for strong models.}
Simulated wrinkles add $+1.5$ yr on average and achieve 19--34\% ACR, with the strongest effects for Custom-Best and VLMs.

\subsection{Combination Attack Results}

Table~\ref{tab:full_combo} reports the full four-attack combination.
Figure~\ref{fig:ridgeplot} (Supplementary) shows the shift in predicted-age distributions from baseline to each attack.

\begin{table}[t]
\centering
\small
\caption{Full combination (Beard + Grey + Makeup + Wrinkles) results. $N_{\text{base}<18}$: baseline minor predictions used as ACR denominator. 95\% Wilson CI computed from $N_{\text{base}<18}$ and ACR.}
\label{tab:full_combo}
\setlength{\tabcolsep}{3.5pt}
\resizebox{\columnwidth}{!}{%
\begin{tabular}{lcrrrr}
\toprule
Model & $N_{\text{base}<18}$ & ACR & 95\% CI & $\Delta\bar{y}$ & MAE$_\text{full}$ \\
\midrule
MiVOLO         &  92 & 72.8\% & [63--81\%] & +6.3 yr & 9.8 \\
Custom-Best    &  77 & 75.3\% & [65--84\%] & +7.2 yr & 9.2 \\
Herosan        &  58 & 67.2\% & [54--78\%] & +11.8 yr & 18.6 \\
MiViaLab       &  68 & 63.2\% & [51--74\%] & +5.9 yr & 11.3 \\
DEX            &  41 & \textbf{82.9\%} & [69--92\%] & +6.8 yr & 12.3 \\
\midrule
Gemini~3 Flash &  80 & 60.0\% & [49--70\%] & +5.9 yr & 8.2 \\
Gemini~2.5 Flash &  85 & 70.6\% & [60--79\%] & +9.3 yr & 13.3 \\
GPT-5-Nano     &  64 & \textbf{59.4\%} & [47--71\%] & +8.5 yr & 11.5 \\
\midrule
\textit{Avg}   & -- & 68.9\% & -- & +7.7 yr & -- \\
\bottomrule
\end{tabular}%
}
\end{table}

\paragraph{The attack converts $\sim$70\% of protected cases.}
Across all models, the four-attack combination achieves an average ACR of 68.9\%: nearly seven in ten images that a model was flagging as minor at baseline are now mispredicted as adult.
The range is 59.4\% (GPT-5-Nano, most robust) to 82.9\% (DEX, most vulnerable).

\paragraph{VLMs show lower ACR.}
The three VLMs achieve ACR of 59--71\%, vs.\ 63--83\% for specialized models.
We note that these ranges overlap (e.g., Herosan CV 67.2\% and MiViaLab CV 63.2\% fall within the VLM range), and no statistical significance test is reported; the observed difference should be interpreted as a trend rather than a definitive finding.
VLMs also predict minor for more subjects in the full-combo subset (64--85 vs.\ 41--92 for specialized), so the absolute number of newly-bypassed cases is comparable even where the rate is lower.

\paragraph{Mean shift of 7.7 years.}
The all-four combination shifts predicted age by $+7.7$ years on average across all 329 subjects (including the 161 young adults aged 18--21 who cannot be bypassed by definition); for the 133 subjects with ground-truth age below 18, the mean shift is $+7.8$ years.
Herosan shows the largest mean shift (+11.8 yr) reflecting its wider prediction variance; Custom-Best shows the best combination of high ACR and large shift.

\subsection{Individual Prediction Trajectories}

\begin{figure*}[t]
  \centering
  \includegraphics[width=\linewidth]{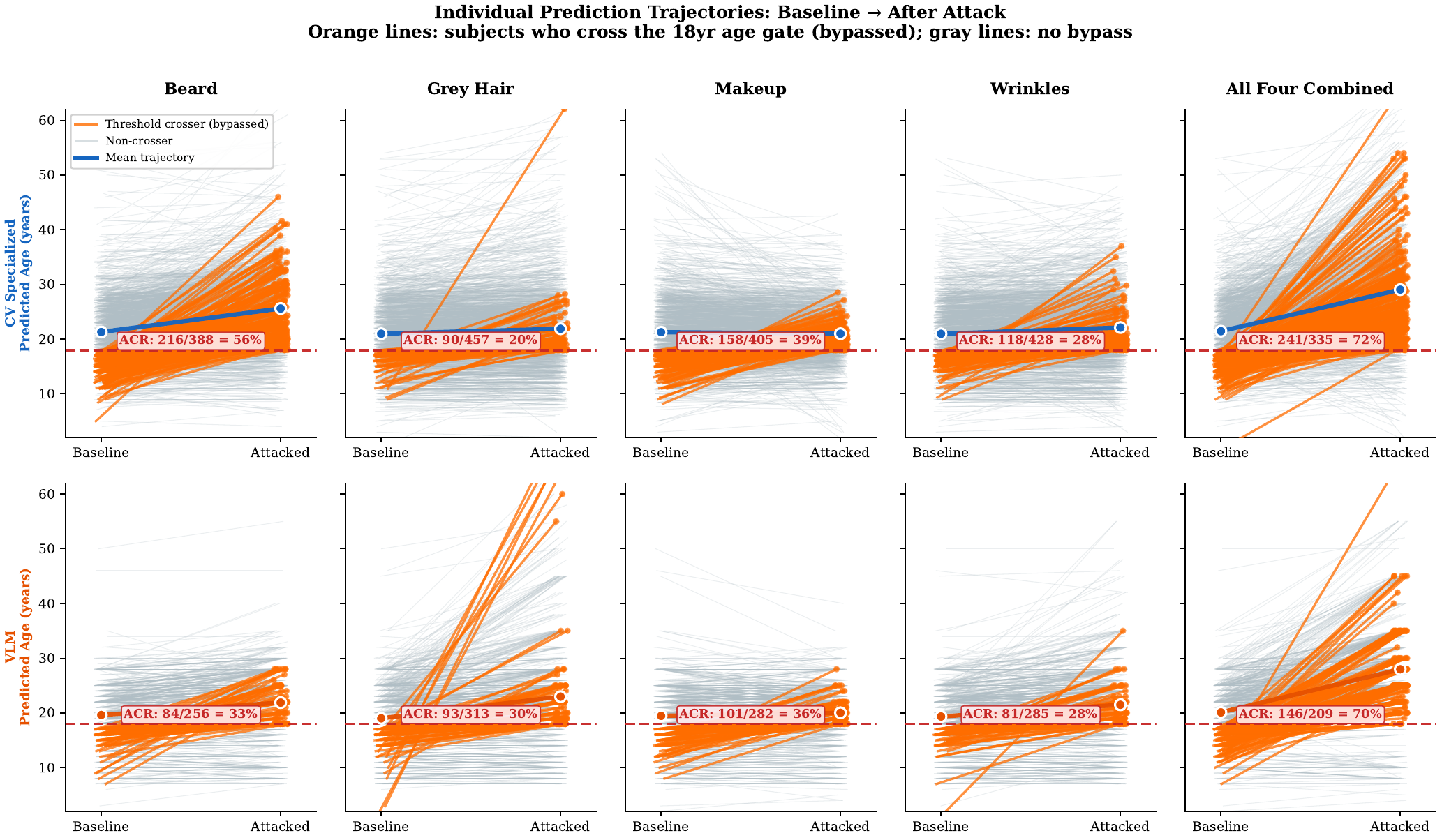}
  \caption{Individual prediction trajectories (baseline $\to$ attacked) for all 329 subjects, per attack type and model class. \textbf{Orange lines}: threshold crossers --- subjects whose predicted age crosses the 18yr gate after attack (these are the events counted by ACR). Gray lines: non-crossers. Bold colored line: group mean trajectory. Red dashed line: 18yr decision threshold. ACR annotation gives the bypass count for each panel.}
  \label{fig:trajectories}
\end{figure*}

Figure~\ref{fig:trajectories} visualizes each subject as a line from their baseline prediction to their attacked prediction, separately for each attack type and model class.
Orange lines mark \emph{threshold crossers}: subjects whose model-assigned age flips from below 18 (flagged as minor) to 18 or above (bypassed as adult) — the exact population that ACR counts.

The trajectory view reveals structure invisible in aggregate statistics.
For the beard attack (CV models), the majority of crossers begin in the 13--17 range and jump to 19--28 — a dramatic shift for a single cosmetic modification.
Makeup shows a distinctive pattern: most subjects' trajectories are nearly horizontal (near-zero shift), but a subset of subjects near the threshold experience large upward jumps, explaining the high ACR despite near-zero mean shift.
The full combination panel shows near-universal upward movement, with crossers spanning all baseline ages up to 17.

\subsection{Age-Stratified Vulnerability}

\begin{figure}[t]
  \centering
  \includegraphics[width=\linewidth]{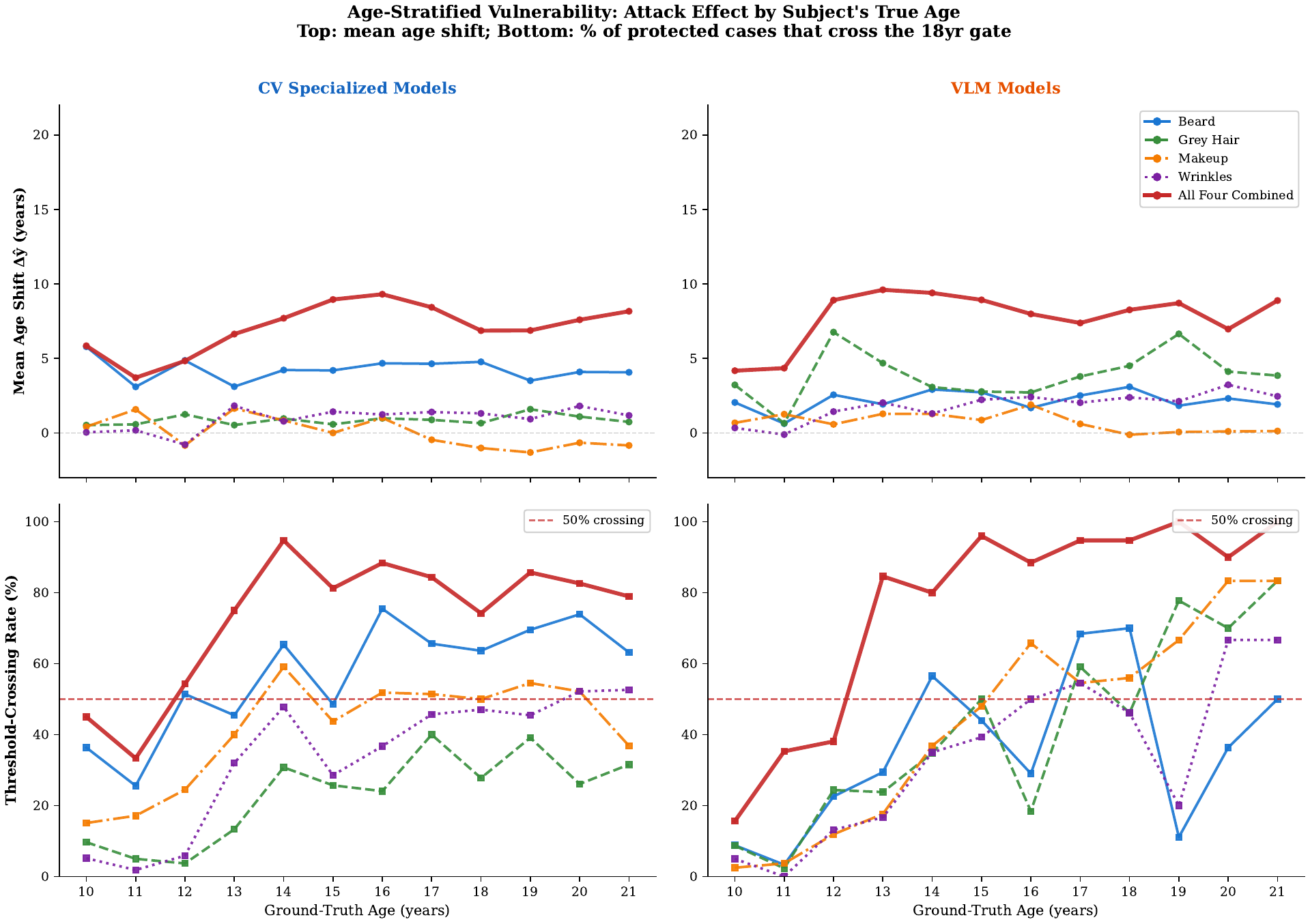}
  \caption{Attack effectiveness stratified by true age. \textbf{Top}: mean age shift per attack type and model class. \textbf{Bottom}: threshold-crossing rate (fraction of model-correctly-identified minors that cross the 18yr gate) per true age. Dashed red line at 50\%. Younger children (ages 10--13) are paradoxically harder to bypass because the model's baseline prediction is further below the threshold; the most vulnerable age group is 15--17 where baseline predictions already cluster near 18.}
  \label{fig:agestratified}
\end{figure}

Figure~\ref{fig:agestratified} disaggregates attack effectiveness by the subject's true (ground-truth) age.
Two complementary metrics are shown: the mean predicted age shift $\Delta\hat{y}$ (top row) and the threshold-crossing rate — the proportion of protected cases converted to adult — per true age group (bottom row).

\paragraph{Younger children are harder to bypass.}
For subjects aged 10--12, the mean baseline prediction is already well below 18, so even a $+5$ yr shift may not cross the threshold.
The threshold-crossing rate is correspondingly low for ages 10--12 across all attacks.

\paragraph{The 15--17 age band is most at risk.}
Subjects aged 15--17 are most vulnerable: their baseline predictions already cluster near the threshold, so even modest cosmetic changes push them over.
This is the most policy-relevant finding: teenagers close to the age boundary face the greatest bypass risk under all attack types.

\paragraph{The combined attack overcomes the age barrier.}
While single attacks (especially beard) plateau for younger children, the all-four combination achieves high crossing rates even for ages 13--14, confirming that attack diversity targets complementary facial cues and provides a more universal bypass.

\subsection{Additive Effect of Combining Attacks}

Both mean shift and ACR grow with the number of combined attacks (from $+1.8$ yr / 34\% ACR averaged over single attacks to $+7.7$ yr / 69\% ACR for the all-four combination), consistent with individual attacks targeting different facial cues with limited overlap.
Mean age shifts for all 15 non-empty subsets are provided in Supplementary Table~\ref{tab:allcombos}; per-model bar charts are in Supplementary Figure~\ref{fig:perbars}.

\section{Conclusion}
\label{sec:conclusion}

We presented the first systematic study of cosmetic attacks on AI age estimation, evaluated across eight diverse models from our prior benchmark.
Using a VLM image editor to simulate physical cosmetic modifications at scale---without any real-world experimentation on human subjects---we showed that simple, household-accessible cosmetics can dramatically compromise age verification systems.

\paragraph{Summary of findings.}
A synthetic beard alone achieves 28--69\% ACR across all eight models, meaning it defeats up to 69\% of cases the model was flagging as minor.
Combining all four attacks (beard, grey hair, makeup, wrinkles) shifts predicted age by $+7.7$ years on average and reaches 83\% ACR for the most vulnerable model.
Even the most accurate specialized model (Custom-Best, MAE 2.71 years) is converted at 75\% ACR under the full attack.
We also introduced ACR as a population-agnostic alternative to simple bypass rates, ensuring results are comparable across test sets with different age distributions.
VLMs tend toward lower ACR (59--71\%) than specialized models (63--83\%) under the full attack, though the ranges overlap and confidence intervals are wide for small-denominator models; whether this trend reflects architectural differences requires further study.

\paragraph{Implications for deployment.}
Our results indicate that any AI-based age gate evaluated solely on clean images will overestimate its real-world effectiveness.
Model selection for age verification should include adversarial robustness as a first-class evaluation criterion.
Specifically: (1)~VLM-based estimators show lower ACR in our evaluation and competitive accuracy, though this advantage should be confirmed with statistical testing at larger scale before treating it as a definitive recommendation; (2)~even the most robust model tested (GPT-5-Nano, 59.4\% ACR) requires complementary controls for dependable age gating; (3)~DEX-style models with coarse age binning should be avoided for age verification, based on their high ACR and low baseline sensitivity to minors (17\% of images predicted minor at baseline).

\paragraph{Limitations and future work.}
Our attacks are simulated, and real-world cosmetics applied under varying lighting, angles, and skin types may show different effects from our digital simulation.
We evaluated only static single-image inputs; systems using video liveness detection or multi-frame reasoning may be more robust.
We studied only four attack types; other modifications (hats, glasses, facial hair styles, lighting manipulation) represent an open threat surface.
The adversary model is passive; active attacks that query the model and optimize the cosmetic configuration represent a stronger threat that merits future study.
Additionally, the beard attack has an inherent gender confound: a beard on a female-presenting face is visually conspicuous and may behave differently from the male-presenting case; future work should include gender-stratified analysis.
We also note that our attack images were generated with Gemini~2.5 Flash Image, which shares the same model family as one of the evaluated victim models (Gemini~2.5 Flash); while these are distinct model endpoints, future work should validate results with an independent image editor.
Furthermore, all attack images originate from a single commercial API; different image editors may produce cosmetic modifications with different visual properties, leading to different ACR values.
VLM image editors are stochastic by nature---the same prompt and image produce different outputs across runs---so the reported ACR values reflect one particular set of generated images; variance across generation runs was not quantified.
We also observe that the Gemini~2.5 Flash image editor disproportionately refused to generate attacked images for minor-presenting subjects: among the 22 images dropped for the beard attack, 86\% were true minors (vs.\ 51\% of the full test set); among the 43 images dropped for the full four-attack combination, 81\% were true minors.
This systematic refusal bias means our ACR estimates may be conservative---the subjects the editor refused are drawn disproportionately from the minor population, so the true bypass rate could be higher if all cases could be evaluated.
Commercial model APIs are also updated continuously without version guarantees, which limits exact reproducibility.
Our mean age shift statistic ($+7.7$ yr) is averaged over all 329 subjects including 161 young adults (ages 18--21) who cannot be bypassed by definition; for true minors (ages 10--17) the mean shift is $+7.8$ yr.
Finally, ACR denominators are small for some models (e.g., DEX: $N=41$), so point estimates carry wide uncertainty; confidence intervals should be reported as evaluations scale to larger datasets.

\paragraph{Data and code availability.}
The evaluation code and per-model prediction results will be released upon publication.
The generated attack images are not redistributed out of an abundance of caution regarding the source dataset licenses; however, all images can be reproduced by applying the prompts in Table~\ref{tab:prompts} to the corresponding source images using Gemini~2.5 Flash Image.
The canonical, maintained locations for the code and non-restricted resources associated with this paper are the Scam.AI Research portal, \url{https://scam.ai/research}~\cite{scamai_research}, and, for document- and PDF-forgery resources, Trustable, \url{https://trustable.ai}~\cite{trustable_ai}; these supersede any prior ad hoc distribution.

\paragraph{Acknowledgments.}
We thank the creators of the UTKFace, IMDB-WIKI, MORPH, AFAD, CACD, FG-NET, APPA-REAL, and AgeDB datasets for making their data publicly available for research.

{
    \small
    \bibliographystyle{ieeetr}
    \bibliography{references}

\begin{thebibliography}{10}

\bibitem{osa2023}
{UK Parliament}, ``Online safety act 2023,'' {\em UK Legislation}, 2023.

\bibitem{eu2022dsa}
{European Parliament}, ``Regulation ({EU}) 2022/2065: Digital services act,''
  {\em Official Journal of the European Union}, 2022.

\bibitem{kosa2022}
{US Senate}, ``Kids online safety act ({KOSA}),'' {\em US Senate Bill S.1409},
  2022.

\bibitem{ofcom2024ageverification}
{Ofcom}, ``Guidance on age verification and age estimation,'' {\em UK
  Communications Regulator}, 2024.

\bibitem{ofcom2023childrenonline}
{Ofcom}, ``Children and parents: Media use and attitudes report 2023.'' Ofcom
  Research Report, 2023.

\bibitem{ren2025deepfakereality}
S.~Ren, H.~Xu, T.~Ng, K.~Zewde, S.~Jiang, R.~Desai, D.~Patil, N.-Y. Cheng,
  Y.~Zhou, and R.~Muthukrishnan, ``Do deepfake detectors work in reality?,''
  {\em arXiv preprint arXiv:2502.10920}, 2025.

\bibitem{ren2026aidetect}
S.~Ren, Y.~Zhou, X.~Shen, K.~Zewde, T.~Duong, G.~Huang, H.~Tiangratanakul,
  T.~Ng, E.~Wei, and J.~Xue, ``How well are open sourced {AI}-generated image
  detection models out-of-the-box: A comprehensive benchmark study,'' {\em
  arXiv preprint arXiv:2602.07814}, 2026.

\bibitem{goodfellow2015explaining}
I.~J. Goodfellow, J.~Shlens, and C.~Szegedy, ``Explaining and harnessing
  adversarial examples,'' {\em ICLR}, 2015.

\bibitem{brown2017adversarial}
T.~B. Brown, D.~Man{\'e}, A.~Roy, M.~Abadi, and J.~Gilmer, ``Adversarial
  patch,'' in {\em NeurIPS Workshop}, 2017.

\bibitem{yin2021advmakeup}
B.~Yin, W.~Wang, T.~Yao, J.~Guo, Z.~Kong, S.~Ding, J.~Li, and C.~Liu,
  ``{Adv-Makeup}: A new imperceptible and transferable attack on face
  recognition,'' in {\em IJCAI}, pp.~1200--1206, 2021.

\bibitem{hu2022sintm}
S.~Hu, X.~Liu, Y.~Zhang, M.~Li, L.~Y. Zhang, H.~Jin, and L.~Wu, ``Protecting
  facial privacy: Generating adversarial identity masks via style-robust makeup
  transfer,'' in {\em CVPR}, pp.~15014--15023, 2022.

\bibitem{shamshad2023clip2protect}
F.~Shamshad, M.~N. Khan, and S.~Khan, ``{CLIP2Protect}: Protecting facial
  privacy using text-guided makeup via adversarial latent code optimization,''
  in {\em CVPR}, 2023.

\bibitem{chen2014cosmetics}
C.~Chen, A.~Dantcheva, and A.~Ross, ``Impact of facial cosmetics on automatic
  gender and age estimation algorithms,'' in {\em International Conference on
  Computer Vision Theory and Applications (VISAPP)}, 2014.

\bibitem{anda2020underage}
F.~Anda, D.~Lillis, N.-A. Le-Khac, and M.~Scanlon, ``Assessing the influencing
  factors on the accuracy of underage facial age estimation,'' {\em arXiv
  preprint arXiv:2012.01179}, 2020.

\bibitem{ren2025benchmark}
S.~Ren, X.~Shen, A.~Raj, A.~Dai, C.~Zhang, Y.~Xu, Z.~Chen, S.~Wu, C.~Gong, and
  Y.~Zhang, ``Out of the box age estimation through facial imagery: A
  comprehensive benchmark of vision-language models vs.\ specialized
  architectures,'' {\em arXiv preprint}, 2025.

\bibitem{guo2009agewild}
G.~Guo, G.~Mu, Y.~Fu, and T.~S. Huang, ``Human age estimation using
  bio-inspired features,'' in {\em CVPR}, pp.~112--119, 2009.

\bibitem{rothe2015dex}
R.~Rothe, R.~Timofte, and L.~Van~Gool, ``{DEX}: Deep expectation of apparent
  age from a single image,'' in {\em ICCVW}, pp.~10--15, 2015.

\bibitem{zhang2019c3ae}
C.~Zhang, S.~Liu, X.~Xu, and C.~Zhu, ``{C3AE}: Exploring the limits of compact
  model for age estimation,'' in {\em CVPR}, pp.~12587--12596, 2019.

\bibitem{kuprashevich2024mivolo}
M.~Kuprashevich and I.~Tolstykh, ``{MiVOLO}: Multi-input transformer for age
  and gender estimation,'' in {\em AIST}, vol.~14486 of {\em LNCS}, Springer,
  2024.

\bibitem{paplhjak2024calltoreflect}
J.~Paplham and V.~Franc, ``A call to reflect on evaluation practices for age
  estimation: Comparative analysis of the state-of-the-art and a unified
  benchmark,'' in {\em CVPR}, 2024.

\bibitem{carlini2017towards}
N.~Carlini and D.~Wagner, ``Towards evaluating the robustness of neural
  networks,'' in {\em IEEE SP}, pp.~39--57, 2017.

\bibitem{sharif2016accessorize}
M.~Sharif, S.~Bhagavatula, L.~Bauer, and M.~K. Reiter, ``Accessorize to a
  crime: Real and stealthy attacks on state-of-the-art face recognition,'' in
  {\em CCS}, pp.~1528--1540, 2016.

\bibitem{sun2024diffam}
Y.~Sun, L.~Yu, H.~Xie, J.~Li, and Y.~Zhang, ``{DiffAM}: Diffusion-based
  adversarial makeup transfer for facial privacy protection,'' in {\em CVPR},
  pp.~9544--9554, 2024.

\bibitem{nist2019fate}
{National Institute of Standards and Technology}, ``{FATE}: Facial age
  estimation technology.'' NIST Technical Report, 2019.

\bibitem{brooks2023instructpix2pix}
T.~Brooks, A.~Holynski, and A.~A. Efros, ``{InstructPix2Pix}: Learning to
  follow image editing instructions,'' in {\em CVPR}, pp.~18392--18402, 2023.

\bibitem{comanici2025gemini25}
G.~Comanici, E.~Bieber, M.~Schaekermann, I.~Pasupat, N.~Sachdeva, I.~Dhillon,
  {\em et~al.}, ``{Gemini 2.5}: Pushing the frontier with advanced reasoning,
  multimodality, long context, and next generation agentic capabilities,'' {\em
  arXiv preprint arXiv:2507.06261}, 2025.

\bibitem{zhang2017utkface}
Z.~Zhang, Y.~Song, and H.~Qi, ``Age progression/regression by conditional
  adversarial autoencoder,'' in {\em CVPR}, pp.~5810--5818, 2017.

\bibitem{wang2018face}
Z.~Wang, X.~Tang, W.~Luo, and S.~Gao, ``Face aging with identity-preserved
  conditional generative adversarial networks,'' in {\em CVPR}, pp.~7939--7947,
  2018.

\bibitem{or2020lifespan}
R.~Or-El, S.~Sengupta, O.~Fried, E.~Shechtman, and I.~Kemelmacher-Shlizerman,
  ``Lifespan age transformation synthesis,'' in {\em ECCV}, pp.~739--755, 2020.

\bibitem{rothe2018imdbwiki}
R.~Rothe, R.~Timofte, and L.~Van~Gool, ``Deep expectation of real and apparent
  age from a single image without facial landmarks,'' {\em IJCV}, vol.~126,
  no.~2-4, pp.~144--157, 2018.

\bibitem{ricanek2006morph}
K.~Ricanek and T.~Tesafaye, ``{MORPH}: A longitudinal image database of normal
  adult age-progression,'' {\em IEEE FG}, pp.~341--345, 2006.

\bibitem{niu2016afad}
Z.~Niu, M.~Zhou, L.~Wang, X.~Gao, and G.~Hua, ``Ordinal regression with
  multiple output {CNN} for age estimation,'' in {\em CVPR}, pp.~4920--4928,
  2016.

\bibitem{chen2015cacd}
B.-C. Chen, C.-S. Chen, and W.~H. Hsu, ``Face recognition and retrieval using
  cross-age reference coding with cross-age celebrity dataset,'' {\em IEEE
  TMM}, vol.~17, no.~6, pp.~804--815, 2015.

\bibitem{panis2016fgnet}
G.~Panis, A.~Lanitis, N.~Tsapatsoulis, and T.~F. Cootes, ``Overview of research
  on facial ageing using the {FG-NET} ageing database,'' {\em IET Biometrics},
  vol.~5, no.~2, pp.~37--46, 2016.

\bibitem{agustsson2017apparent}
E.~Agustsson, R.~Timofte, S.~Escalera, X.~Baro, I.~Guyon, and R.~Rothe,
  ``Apparent and real age estimation in still images with deep residual
  regressors on {APPA-REAL} database,'' in {\em IEEE FG}, pp.~87--94, 2017.

\bibitem{moschoglou2017agedb}
S.~Moschoglou, A.~Papaioannou, C.~Sagonas, J.~Deng, I.~Kotsia, and
  S.~Zafeiriou, ``{AgeDB}: The first manually collected, in-the-wild age
  database,'' in {\em CVPRW}, pp.~51--59, 2017.

\bibitem{google2025gemini3flash}
{Google DeepMind}, ``{Gemini 3 Flash} model card,'' tech. rep., Google
  DeepMind, 2025.

\bibitem{openai2025gpt5}
{OpenAI}, ``{OpenAI GPT-5} system card,'' {\em arXiv preprint
  arXiv:2601.03267}, 2025.

\bibitem{scamai_research}
{Scam.AI}, ``Scam.AI Research: Deepfake detection research and datasets,'' 2026.
  [Online]. Available: \url{https://scam.ai/research}.

\bibitem{trustable_ai}
{Trustable}, ``Trustable: Document and {PDF} forgery detection,'' 2026. [Online].
  Available: \url{https://trustable.ai}.

\end{thebibliography}
}

\appendix

\section{Supplementary Figures}
\label{sec:appendix}

\begin{figure*}[t]
  \centering
  \includegraphics[width=\linewidth]{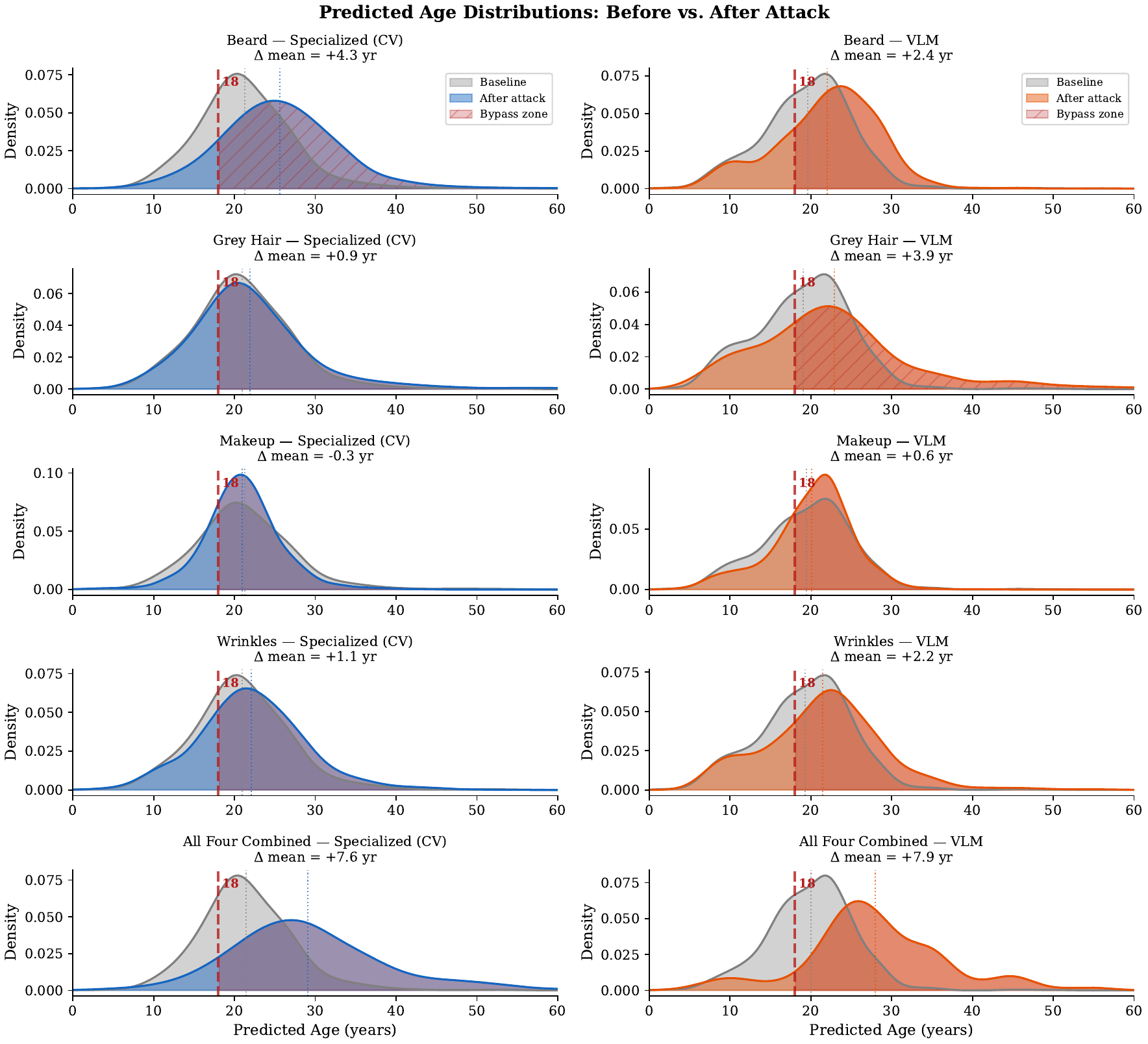}
  \caption{Predicted age distributions before (grey) and after (colored) each attack, separately for CV-specialized and VLM models. Dashed red line = 18yr decision threshold; hatched region = bypass zone (predicted adult). VLM distributions shift rightward more uniformly; CV models show sharper concentration near the gate.}
  \label{fig:ridgeplot}
\end{figure*}

\begin{figure*}[t]
  \centering
  \includegraphics[width=\linewidth]{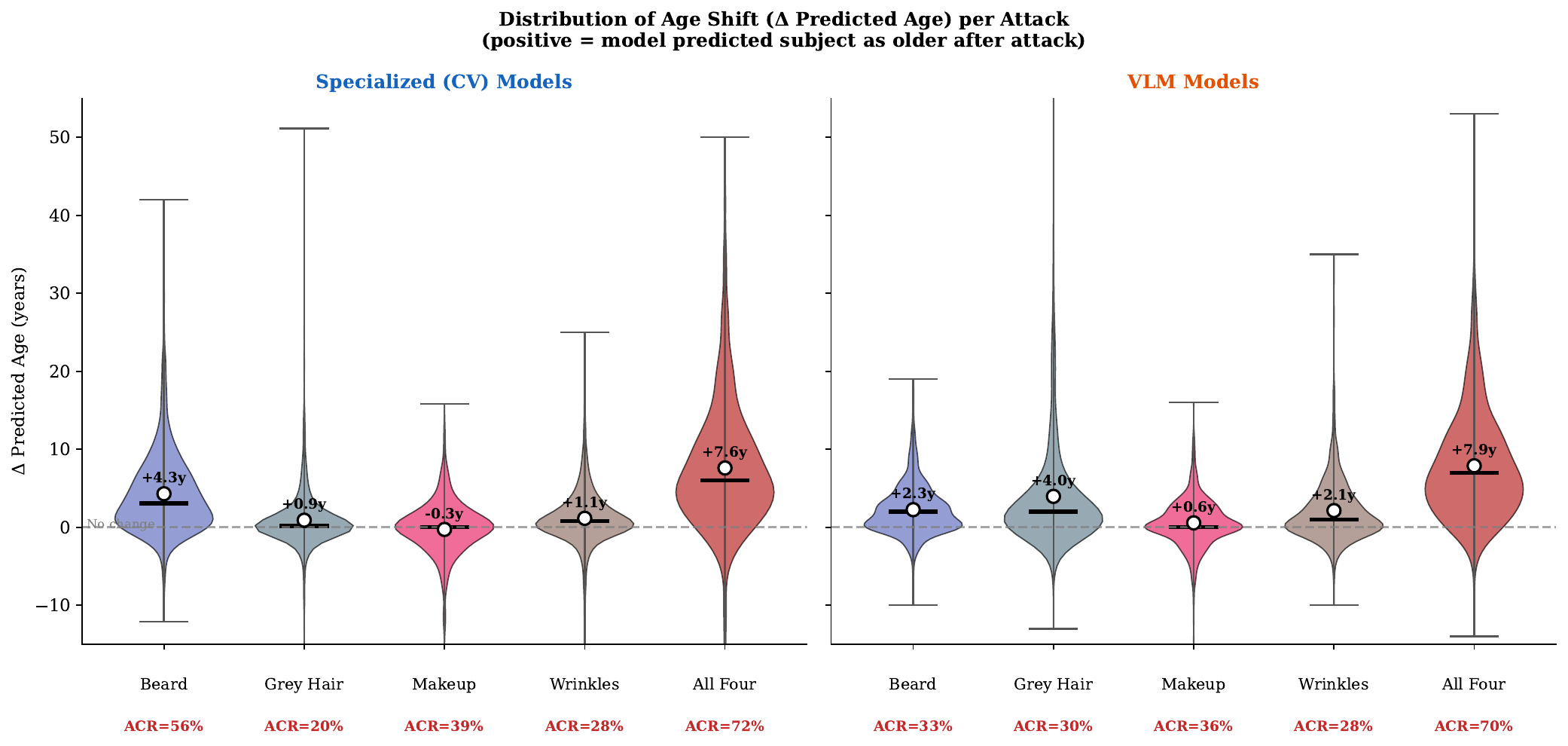}
  \caption{Distribution of individual age shifts ($\Delta$ predicted age) per attack, separated by model type. White dot = mean; ACR = Attack Conversion Rate (below x-axis). The beard attack shows the widest and most skewed distribution for specialized models; grey hair is more effective for VLMs. Makeup exhibits a distinctive bimodal distribution: most subjects shift by $\approx$0 yr, while a subgroup near the threshold shifts upward substantially.}
  \label{fig:violin}
\end{figure*}

\begin{figure*}[t]
  \centering
  \includegraphics[width=\linewidth]{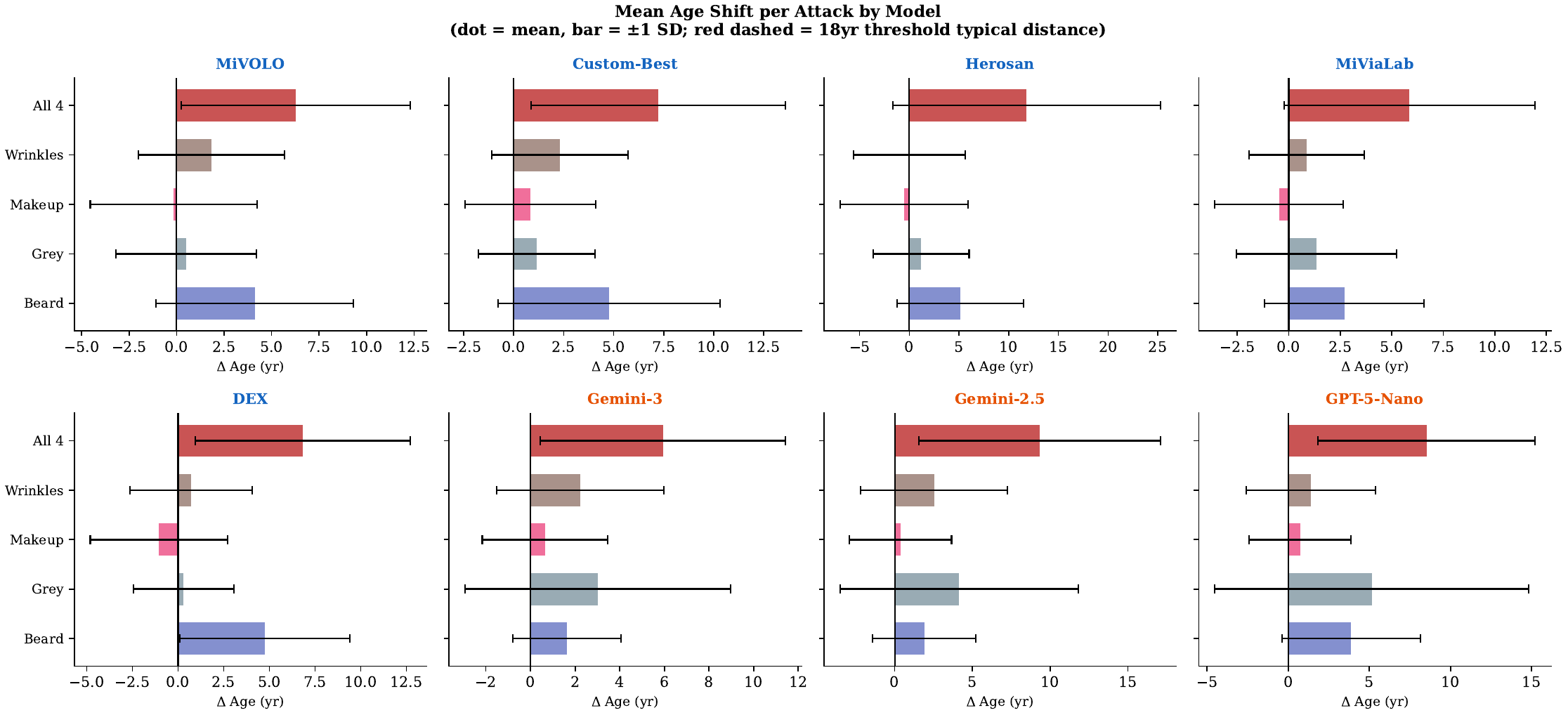}
  \caption{Mean age shift ($\pm$1 SD) per attack and per model. Beard dominates for CV-specialized models; grey hair is comparably effective for VLMs. The all-four combination consistently produces the largest shifts across all models.}
  \label{fig:perbars}
\end{figure*}

\begin{table*}[t]
\centering
\small
\caption{Mean age shift (years) for all 15 non-empty subsets of the four cosmetic attacks, across all 8 models. Positive values indicate the model predicts the subject as older after attack. Bold = row maximum. CV-specialized models separated from VLMs by vertical rule.}
\label{tab:allcombos}
\setlength{\tabcolsep}{4pt}
\resizebox{\textwidth}{!}{%
% Appendix table: mean age shift (years) for all 15 attack combinations
% Positive = attacked image predicted older; bold = row maximum
\begin{tabular}{l *{5}{r} | *{3}{r}}
\toprule
\textbf{Attack Combination} & \textbf{MiVOLO} & \textbf{Custom-Best} & \textbf{Herosan} & \textbf{MiViaLab} & \textbf{DEX} & \textbf{Gemini-3 Flash} & \textbf{Gemini-2.5 Flash} & \textbf{GPT-5-Nano} \\
\midrule
Beard & +4.1 & +4.8 & \textbf{+5.1} & +2.7 & +4.8 & +1.6 & +1.9 & +3.9 \\
Grey Hair & +0.5 & +1.2 & +1.2 & +1.4 & +0.3 & +3.0 & +4.1 & \textbf{+5.2} \\
Makeup & -0.1 & \textbf{+0.8} & -0.6 & -0.5 & -1.1 & +0.6 & +0.4 & +0.7 \\
Wrinkles & +1.8 & +2.3 & -0.0 & +0.9 & +0.7 & +2.2 & \textbf{+2.5} & +1.4 \\
\midrule
Beard + Grey & +5.1 & +6.1 & +8.6 & +4.7 & +5.0 & +4.7 & +7.5 & \textbf{+9.2} \\
Beard + Makeup & +3.9 & +5.3 & \textbf{+5.6} & +2.9 & +4.5 & +2.2 & +3.2 & +4.7 \\
Beard + Wrinkles & +5.5 & +6.5 & \textbf{+6.5} & +4.0 & +5.5 & +3.5 & +4.6 & +5.4 \\
Grey + Makeup & +1.6 & +2.1 & +2.5 & +1.4 & +0.7 & +3.1 & \textbf{+5.0} & +4.3 \\
Grey + Wrinkles & +2.1 & +3.0 & +3.5 & +2.6 & +1.9 & +5.0 & \textbf{+7.0} & +5.8 \\
Makeup + Wrinkles & +2.8 & \textbf{+3.0} & +1.7 & +1.4 & +1.7 & +2.1 & +2.6 & +1.3 \\
\midrule
Beard + Grey + Makeup & +4.6 & +5.8 & \textbf{+8.9} & +4.3 & +4.9 & +4.4 & +7.2 & +7.4 \\
Beard + Grey + Wrinkles & +5.9 & +7.1 & \textbf{+10.6} & +5.8 & +6.3 & +6.2 & +9.3 & +9.7 \\
Beard + Makeup + Wrinkles & +6.1 & +6.7 & \textbf{+7.7} & +4.4 & +6.1 & +3.7 & +4.7 & +5.7 \\
Grey + Makeup + Wrinkles & +3.1 & +3.6 & +5.8 & +2.6 & +2.6 & +4.6 & \textbf{+7.1} & +5.4 \\
\midrule
All Four & +6.3 & +7.2 & \textbf{+11.8} & +5.9 & +6.8 & +5.9 & +9.3 & +8.5 \\
\bottomrule
\end{tabular}%
}
\end{table*}

\end{document}